\documentclass[twoside,11pt]{article}

\usepackage[arxiv]{melba}
\usepackage{mwe} 

\usepackage{amsmath,amsfonts}
\usepackage{algorithm}
\usepackage{algpseudocode}

\melbaid{2024:022}  
\doi{https://doi.org/10.59275/j.melba.2024-f5fc}
\melbaauthors{Buddenkotte et. al.}  
\volume{2}
\firstpageno{2067}  
\melbayear{2024}  
\datesubmitted{08/2024}  
\datepublished{10/2024}  

\melbaspecialissue{Medical Imaging with Deep Learning (MIDL) 2020}
\melbaspecialissueeditors{Marleen de Bruijne, Tal Arbel, Ismail Ben Ayed, Hervé Lombaert}

\ShortHeadings{CTARR}{Buddenkotte et. al.}

\title{CTARR: A fast and robust method for identifying anatomical regions on CT images via atlas registration}

\author{\firstname Thomas \surname Buddenkotte \email thomasbuddenkotte@googlemail.com \\  
	\addr Department of Diagnostic and Interventional Radiology and Nuclear Medicine, University Medical Center Hamburg-Eppendorf, Hamburg, Germany,\\
    jung diagnostics, Hamburg Germany \\
    Department of Oncology, University of Cambridge, UK\\
    \AND
    \name Roland Opfer
    \addr jung diagnostics, Hamburg Germany \\
    \AND
    \name Julia Kr{\"u}ger
    \addr jung diagnostics, Hamburg Germany \\
    \AND
    \name Alessa Hering
    \addr Department of Medical Imaging, Radboud University Medical Center, Nijmegen, The Netherlands \\
    \AND
    \name Mireia Crispin-Ortuzar
    \addr Early Cancer Institute, Department of Oncology, University of Cambridge, UK\\
    Cancer Research UK Cambridge Centre, University of Cambridge, UK\\
}

\begin{document}

\maketitle

\begin{abstract}
    Medical image analysis tasks often focus on regions or structures located in a particular location within the patient's body. Often large parts of the image may not be of interest for the image analysis task. When using deep-learning based approaches, this causes an unnecessary increases the computational burden during inference and raises the chance of errors. In this paper, we introduce CTARR, a novel generic method for CT Anatomical Region Recognition. The method serves as a pre-processing step for any deep learning-based CT image analysis pipeline by automatically identifying the pre-defined anatomical region that is relevant for the follow-up task and removing the rest. It can be used in (i) image segmentation to prevent false positives in anatomically implausible regions and speeding up the inference, (ii) image classification to produce image crops that are consistent in their anatomical context, and (iii) image registration by serving as a fast pre-registration step. Our proposed method is based on atlas registration and provides a fast and robust way to crop any anatomical region encoded as one or multiple bounding box(es) from any unlabeled CT scan of the brain, chest, abdomen and/or pelvis. We demonstrate the utility and robustness of the proposed method in the context of medical image segmentation by evaluating it on six datasets of public segmentation challenges. The foreground voxels in the regions of interest are preserved in the vast majority of cases and tasks (97.45-100\%) while taking only fractions of a seconds to compute (0.1-0.21s) on a deep learning workstation and greatly reducing the segmentation runtime (2.0-12.7x). Our code is available at https://github.com/ThomasBudd/ctarr.
\end{abstract}


\section{Introduction}
Deep learning is currently the most dominant technology for automated medical image analysis tasks such as classification, segmentation and prognosis. Computed tomography (CT) is one of the most common medical imaging modalities nowadays (\cite{CT}\footnote{https://www.who.int/data/gho/data/indicators/indicator-details/GHO/total-density-per-million-population-computed-tomography-units}\footnote{https://www.england.nhs.uk/statistics/wp-content/uploads/sites/2/2022/07/Statistical-Release-21st-July-2022-PDF-875KB.pdf}). The field of view CT images varies naturally. For example, on a large scale the medical question determines which body part is scanned (head, chest, abdomen, pelvis etc.). On a small scale it depends on the input of the radiographers to the scanner and thus slightly varies between users. On the other hand, the information required for the medical question or the image analysis task is often located in a particular anatomical region of interest. \cite{kits23_winner} suggested a segmentation pipeline build on deep supervision, a method where a preliminary low resolution segmentation is created on the entire scan and a full-resolution network refines the prediction only in the context around the preliminary segmentation. \cite{sybil_lung_cancer} proposed an image classification pipeline to assess lung cancer risk that used automated segmentation of the lung and applied the classification network only inside a crop around this segmentation. These and similar approaches heavily rely on a robust segmentation of the organ in which the disease is located and thus cannot be applied directly to other CT image analysis problems, such as segmentation of metastatic disease, where lesions can appear in a variety of locations. Further, computing bounding boxes based on any automated segmentations can be disadvantageous for several other reasons. First, such automated segmentation algorithms tend to produce false positive annotations far away from the actual anatomical region of interest as previously reported by \cite{ovarian_segmentation} and demonstrated in Appendix A. Second, predicting the segmentation on full volume on full resolution of the CT image can be computationally expensive. Third, using a segmentation algorithm on low resolution instead is not feasible in problems where the object of interest is very small such as lymph nodes or bone fractures.\\
In this manuscript, we describe a fast and robust approach for the identification of predefined anatomical regions in CT scans. In contrast to the previously described approaches by \cite{kits23_winner} and \cite{sybil_lung_cancer}, our method can identify any anatomical region of interest without the need for fine-tuning the pipeline. Instead, new anatomical regions of interest are simply added as bounding boxes in an atlas coordinate system. The method uses image registration to map such bounding boxes from the atlas coordinate system to the coordinate system of the incoming CT scan. We further suggest a novel image registration algorithm that is particularly suited for this task. Traditional registration methods often use iterative schemes and are likely to get stuck in local minima in cases where large translations are needed for the optimal alignment. Our method prevents this by using anatomical segmentation masks instead of the CT images directly. We validate our method on a total of 1131 CT scans from public segmentation challenges to demonstrate robustness and computational feasibility. The proposed method also identifies orientation misalignment compared to the atlas with regards to 90-, 180- and 270-degrees' rotations in xy-plane and reversion of the z-axis, which occur in some dicom to nifti conversion tools or when false information was added into the dicom header. The proposed method can serve as a pre-processing step for CT image analysis problems like image segmentation, classification or registration. The framework is available for at https://github.com/ThomasBudd/ctarr and offers a convenient interface to crop from predefined anatomical regions on unlabeled scans as well as inferring custom ones from labelled datasets.

\section{Related Works}

To the best of our knowledge, our method is the only one today that can extract any anatomical region of interest from a CT scan as previous approaches are tailored for only a single region. Previous approaches are typically build on either image registration or segmentation while our method utilises a combination of both. In the following we will describe these previously existing method.
\subsection{Image registration-based approaches}
There is a long tradition in creating specialised anatomical region recognition in medical images that use image registration. For example, \cite{atlas_registration_segmentation_survey} summarized well before the age of deep learning existing techniques to perform image segmentation by registering an image to an atlas and propagating the segmentation from the atlas to the image. The usage of atlas-based registration techniques is especially popular in brain imaging. CT, MRI, or SPECT images are often registered to an atlas in the so-called Montreal Neuro Imaging (MNI) space during pre-processing before carrying out further analysis such as accurate identification of different brain regions as suggested by \cite{CerebrA_MNI} or \cite{MNI_reg}. Similar to our approach, \cite{ralph_spect} created a registration method for brain SPECT images to the MNI space to consistently crop from six slices with locations defined in this MNI space. The reason why these approaches are popular in the field of brain imaging might be that the robust and reliable registration algorithms could be proposed due to the rigidity of the brain. Image registration in other areas of the body can be significantly more challenging, especially considering anatomies that are more flexible or show more variation between patients. Nowadays, deep learning is often used as a part of image registration pipeline to solve difficult registration problems or decrease the computational effort of conventional approaches. For a comprehensive survey we refer to \cite{im_reg_survey_2021} and \cite{im_reg_survey_2023}.\\
\subsection{Image segmentation-based approaches}
Other approaches comparable to our proposed method automatically segment organs of interest to create smaller image crops. \cite{sybil_lung_cancer} suggested a prediction pipeline for lung cancer risk and used cropping around automatically created lung segmentations as a prepossessing step. \cite{kits23_winner} won the 2023 kidney tumor segmentation challenge (kits23)\footnote{https://kits-challenge.org/kits23/} by using an approach that first identifies the kidneys on a low resolution, cropping around this region and employing a second network that operates only on the full resolution crops.\\
Our method is based on anatomy segmentation of the full body and registration to an atlas which are active fields of research. One of the most relevant current work on segmentation of various anatomies on CT images is the Totalsegmentator by \cite{totalsegmentator}. The Totalsegmentator consists of a dataset of more than 1200 CT scans and annotations from 117\footnote{The original version had annotations from 104 anatomies while the recent update contains 117 https://github.com/wasserth/TotalSegmentator} different anatomies and automated segmentation models for the segmentation for those anatomies. The annotations were made by first creating a small set of manual annotations, training a publicly available segmentation model on these and refining the model predictions on remaining scans. Other examples of automated segmentation of organs and other anatomies can be found in the well-established nnU-Net framework \cite{nnunet}, where the authors proposed a pipeline that automatically suggests hyper-parameters for a CNN-based segmentation model given any new dataset for biomedical image segmentation. The framework won multiple segmentation challenges and can be downloaded freely with pre-trained weights for multiple segmentation tasks including organ segmentation. For example, the challenge submission for the Multi Atlas Labeling Beyond the Cranial Vault - Abdomen Challenge is capable of segmenting 13 different organs in the abdomen. For a more comprehensive review on anatomy segmentation on CT images can be found in \cite{totalsegmentator}\\
\subsection{Comparison to our method}
All previously established approaches have in common that they are specialised for one particular anatomical region of interest. Modification to other anatomical regions often involves collecting new datasets, training new networks or adapting parameters in the registration pipeline. In contrast to this, our method is a general-purpose tool in the sense that new anatomical regions can be added by defining a new bounding box in the coordinate system of an atlas. The rest of the pipeline is agnostic to the choice of the bounding box. Our novel image registration method uses sets of segmentation masks instead of performing on the CT images directly, this allows us to prevent the method from getting caught in local minima and to safely determine even large transformations as we will demonstrate in the following sections.

\section{Material and Methods}

In the following we will describe the proposed method in detail. The goal of our method is to automatically identify pre-defined anatomical regions in any incoming CT scan. This is achieved by encoding anatomical regions as one or multiple bounding boxes in the coordinate system of the atlas and mapping those to the CT scan by performing image registration. In contrast to many existing image registration techniques, we do not perform the image registration directly on the CT images, but instead on the segmentation masks of a fixed set of 19 anatomies. We will first describe our automated segmentation method to obtain this segmentation automatically in Section \ref{anatomy_segmentation} followed by the atlas registration method in Section \ref{atlas_registration}. The pipeline of how to perform the cropping of a pre-defined anatomical region on an unseen CT image is described in Section \ref{region_cropping}. Section \ref{region_computing} described how such anatomical region can be computed when none is pre-defined, but a set of CT images and segmentations of the region of interest are given.

\subsection{Segmentation of anatomical structures}\label{anatomy_segmentation}
The segmentation of the anatomical structures was performed by training a 3d U-Net (\cite{unet2d, unet3d}) on the Totalsegmenator dataset. We first carefully selected a set of 19 anatomical structures throughout the whole body which ratified the following criteria: (1) large volume, (2) segmented by previous approaches with high accuracy and (3) do not demonstrate large anatomical variations. We further created groups of some anatomies to reduce the complexity of the segmentation task. The resulting list of the 19 anatomical structures considered by our segmentation approach can be found in Table \ref{tab:anatomy_list}. We ensured that our segmentation model is on state-of-the-art level by starting from hyper-parameters suggested by the nnU-Net framework followed by a hyper-parameter tuning of the patch size, learning rate, batch size, optimizer, and augmentation strength. 
\begin{table}[h] 
    \centering
    \caption{Target anatomies of the segmentation approach}
    \begin{tabular}{l | l || l | l}
        \textbf{ID} & \textbf{Anatomy name} & \textbf{ID} & \textbf{Anatomy name} \\
        \hline
        1 & skull & 11 & brain\\ 
        2 & C vertebrae & 12 & lung left\\
        3 & left rips, scapula, clavicular & 13 & lung right\\
        4 & right rips, scapula, clavicular & 14 & heart\\
        5 & sternum & 15 & liver \\
        6 & T vertebrae & 16 & spleen\\
        7 & L vertebrae & 17 & left kidney\\
        8 & sacrum & 18 & right kidney\\
        9 & left hip & 19 & urinary bladder\\
        10 & right hip &&\\        
    \end{tabular}
    \label{tab:anatomy_list}
\end{table}
For pre-processing, we resized the training dataset to 3mm isotropic voxel spacing and created a three channel input by windowing the CT image with a bone, lung and soft-tissue window followed by normalizing the gray values to [0, 1] in each channel. The network is a standard 3d U-Net with 32 filters in the first block, four stages and 20 output channels followed by a softmax function. The network was trained using the ADAMW optimizer as suggested by \cite{adamw} for 250.000 steps with a batch size of 4, a linear warm-up plus cosine decay schedule with a maximum learning rate of 0.0016, $\beta_1=0.98$, $\beta_2=0.999$ and a weight decay of 0.0001. We used a cubed input patches of $64^3$ voxels. To promote robustness of the network, we employed aggressive data augmentation during training, namely rotation in xy plane, z axis flipping, zooming in and out as well as heavy Gaussian noise and blurring (see Supplementary Materials for more Detail). In contrast to other segmentation frameworks like nnU-Net by \cite{nnunet}, we do not resize the segmentations to the original resolution, but instead perform the atlas registration step using the isotropic resolution of 3mm to reduce the memory and computational cost.\\
Recent research of \cite{nnunetrevisited} suggests that the performance of simple 3d U-Nets can be improved by simply changing the decoder to a ResNet and increasing the amount of convolutional layers. However, this comes at the expense of increased computational burden. To prevent this and maintain a high inference speed we decided against more elaborate architectures. Instead, we research how improved segmentation quality affects the registration results by applying the registration once with our automated segmentations and once with ground truth segmentations.

\subsection{Atlas registration}\label{atlas_registration}
The aim of the atlas registration is to find a mapping $T$ which aligns the input CT image $im$ (moving image) to the atlas (fixed image). Classical methods use the CT images directly and have the disadvantage of being sensitive towards the initial condition of the iterative scheme in the sense that they often get stuck in local minima in cases where large transformations are needed to align the two images. The registration method presented in this prevents this by \textit{not} acting on CT images directly, but instead using the output of the anatomy segmentation described in the previous section. It is important to note that the atlas is \textit{not} present as a CT image, but as a set of averaged segmentations of the anatomies listed in Table \ref{tab:anatomy_list}. The creation of this atlas of segmentations is described in the end of this section.\\
In principle any type of registration algorithm can be used to map the bounding box(es) from the atlas coordinate system to the scans coordinate system. In this work we decided to restrict the transformation such that the topology of bounding boxes is maintained after transformation. To be precises, we restrict the transformation such that transformed bounding boxes still have edges parallel to the x-, y-, and z-axis by allowing only translations, scalings and possibly 90-, 180- and 270-degrees rotations in the xy-plane as well as flippings along the z-axis. The computation of this transformation is divided in two steps, a first translation alignment followed by an iterative gradient descent step.\\
The first translational alignment step prevents the iterative scheme from getting stuck in local minima, especially for cases where large translations are needed for optimal alignment. To compute this translation, we compute the center of masses of the segmentations and used them as landmarks to minimize a weighted mean squared error. For this, let $x\in\mathbb{R}^{19\times3}$ be the center of mass of the segmentations from the input CT image and $y\in\mathbb{R}^{19\times3}$ be the center of mass of the segmentations of the atlas. The coordinates in $x$ are only well-defined for anatomies that are covered in the input CT image. To put no weight on anatomies not covered in the input image and only little weight on those only partially covered, we introduce a weighting vector $w$. This weighting vector is computed on the fly for each incoming image. For anatomy $i$, $\tilde{w}_i$ is computed as the ratio volume in the input image vs the volume of this anatomy in the atlas. The resulting vector is normalized to sum to 1 ($w_i=\tilde{w}_i/\sum_j\tilde{w}_j$). Given the two sets of center of masses $x$ and $y$ and a weighting vector $w$, we obtain the translation via
\begin{align*}
    t_j &= \text{argmin} \sum_{i=1}^{19} w_i (x_{ij} + t_j - y_{ij})^2\\
    &= \sum_{i=1}^{19}w_i (y_{ij} - x_{ij})
\end{align*}
To account for misalignment of the atlas and input image in terms of 90-, 180-, or 270-degree rotations in the xy plane or a flipped z-axis, we iterate over all possible configurations and choose the one that minimizes the previously described weighted mean squared error of the landmarks. The landmark-based registration is computationally cheap, but contains imperfections as the landmarks contain less information than the set of segmentation masks. The initial translational alignment is a crucial step as it prevents the iterative refinement scheme from getting stuck in local minima like traditional image-based methods do.\\
To improve upon the landmark-based registration, we perform a gradient descent-based refinement scheme to obtain the final translation and scaling component of the registration transformation. This is performed by minimizing the dice loss between the segmentation masks of the CT image $seg_{im}$ and the atlas $seg_{atlas}$
\[ \min_T L^{dsc}(T(seg_{im}), seg_{atlas}),\]
where $T$ denotes the registration operator that maps the moving image to the fixed.
We initialise the parameters of the translation as the ones computed by the landmark-based approach and the parameters of the scaling as ones (no scaling). Next, we apply gradient descent to minimize the dice loss between the anatomy segmentations of the input CT image (moving image) and the atlas (fixed image). We use an initial step size of 0.05 and reduce the step size by a factor of 2 each time the loss did not decrease by at least 0.005 between two iterations. The iteration is stopped when reducing the step size a fourth time. The rotations in xy-plane the and flipping of the z-axis are left unchanged in this stage.\\
The atlas of anatomical segmentations was obtained by registering and averaging all 37 scans from the Totalsegmentator that contained all 19 segmentation classes. First, each of these scans was registered to all remaining 36 scans and the average over all 36 translations and scalings was computed. Second, these averaged transformations were applied to each corresponding set of segmentations of the anatomical structures. Third, we computed a provisionally atlas by computing the voxel-vise average of the transformed anatomy segmentations. Fourth, each scan was registered to the provisionally atlas. The final atlas was computed by the voxel-wise average of the aforementioned registered anatomy segmentations and checked visually.

\subsection{Cropping of pre-defined anatomical regions}\label{region_cropping}

\begin{figure}[h]
    \centering
    \includegraphics[width=1.0\linewidth]{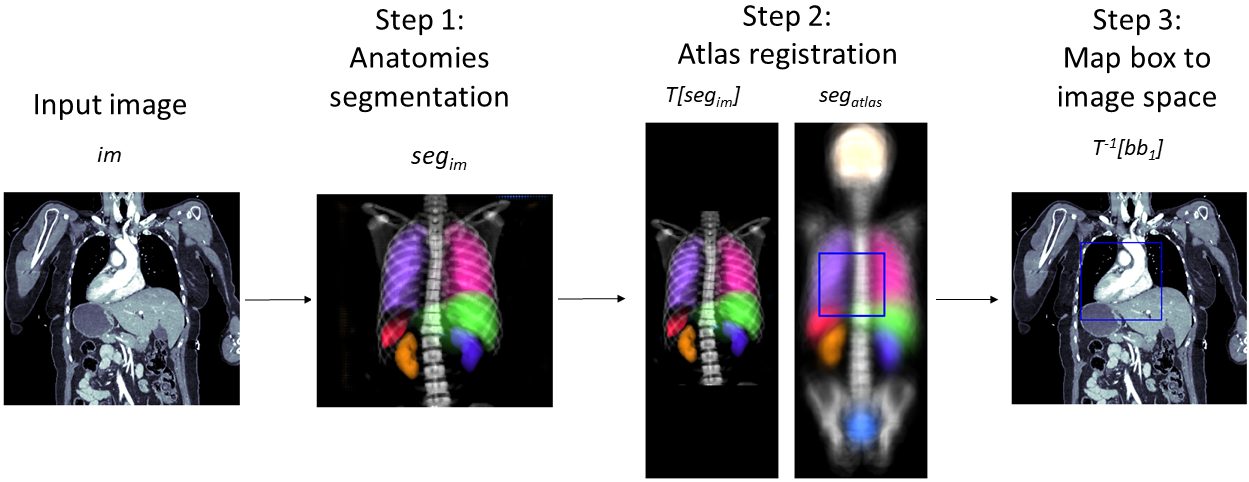}
    \caption{Pipeline of cropping pre-computed anatomical regions in unseen CT images. Step 1: CNN-based segmentation for a fixed set of 19 anatomies is performed. Step 2: image registration is performed to align the segmentations of step 1 (moving image) to the atlas (fixed image). Step 3: the registration transformation is being used to map the anatomical region from the atlas- to the image coordinate system and used for cropping. The segmentations are displayed as colored pseudo X-Ray images where the sum of the binary mask was computed over the x-axis. }
    \label{fig:cropping_pipeline}
\end{figure}

Let's assume that the anatomical region of interest is given as one or multiple bounding boxes $bb_1,\dots,bb_k$ in the coordinate system of the atlas. To map those bounding box(es) to the coordinate system of the input image, one first has to compute the registration transformation $T$ as described in Section \ref{atlas_registration}. As $T$ maps the input image (moving image) to the atlas (fixed image), $T^{-1}$ maps the bounding boxes $bb_1,\dots,bb_k$ to the input images coordinate system. By design of $T$ the mapped bounding boxes $T^{-1}(bb_1),\dots,T^{-1}(bb_k)$ still have edges parallel to the x-, y-, and z-axis and thus can be used for cropping without any need for interpolation. This workflow is visualized in Figure \ref{fig:cropping_pipeline} and presented as pseudo code in Algorithm \ref{alg:cropping}. 

\begin{algorithm}
\caption{CT Anatomical Region Cropping}\label{alg:cropping}
\begin{algorithmic}
\Require $im,\:seg_{atlas},\:bb$ \Comment{CT image, atlas segmentation, bounding box}
\State $seg_{im} \gets \text{AnatomySegmentation}(im)$
\State $x \gets \text{CenterOfMass}(seg_{im})$ \Comment{Landmarks moving image}
\State $y \gets \text{CenterOfMass}(seg_{atlas})$ \Comment{Landmarks fixed image}
\State $\tilde{w} \gets \text{sum}(seg_{im}, (1,2,3)) /\text{sum}(seg_{atlas}, (1,2,3))$
\State $w \gets \tilde{w} / sum(\tilde{w})$ \Comment{MSE Weight}
\State $t_0,\:k_{rot},\:f_{z} \gets \text{LeastSquares}(x,\:, y\:, w)$ \Comment{Orientation and initial translation}
\If{$k_{rot}>0$}
\State $seg_{im} \gets \text{Rotate90}(seg_{im}, k_{rot})$ \Comment{xy rotation}
\State $im \gets \text{Rotate90}(im, k_{rot})$
\EndIf
\If{$f_{z}$}
\State $seg_{im} \gets \text{FlipZ}(seg_{im})$ \Comment{Flipping of z axis}
\State $im \gets \text{FlipZ}(im)$
\EndIf
\State $s_1,\:t_{1}\gets \text{IterativeRegistration}(seg_{im}, seg_{atlas}, t_0)$ \Comment{Refine registration}
\State $bb_{im} \gets (bb - t_1)/s1$ \Comment{Map bounding box to scan}
\State $im_{crop} \gets \text{Cropping}(im,\: bb_{im})$
\end{algorithmic}
\end{algorithm}

\subsection{Computing of new anatomical regions}\label{region_computing}

\begin{figure}[h]
    \centering
    \includegraphics[width=1.0\linewidth]{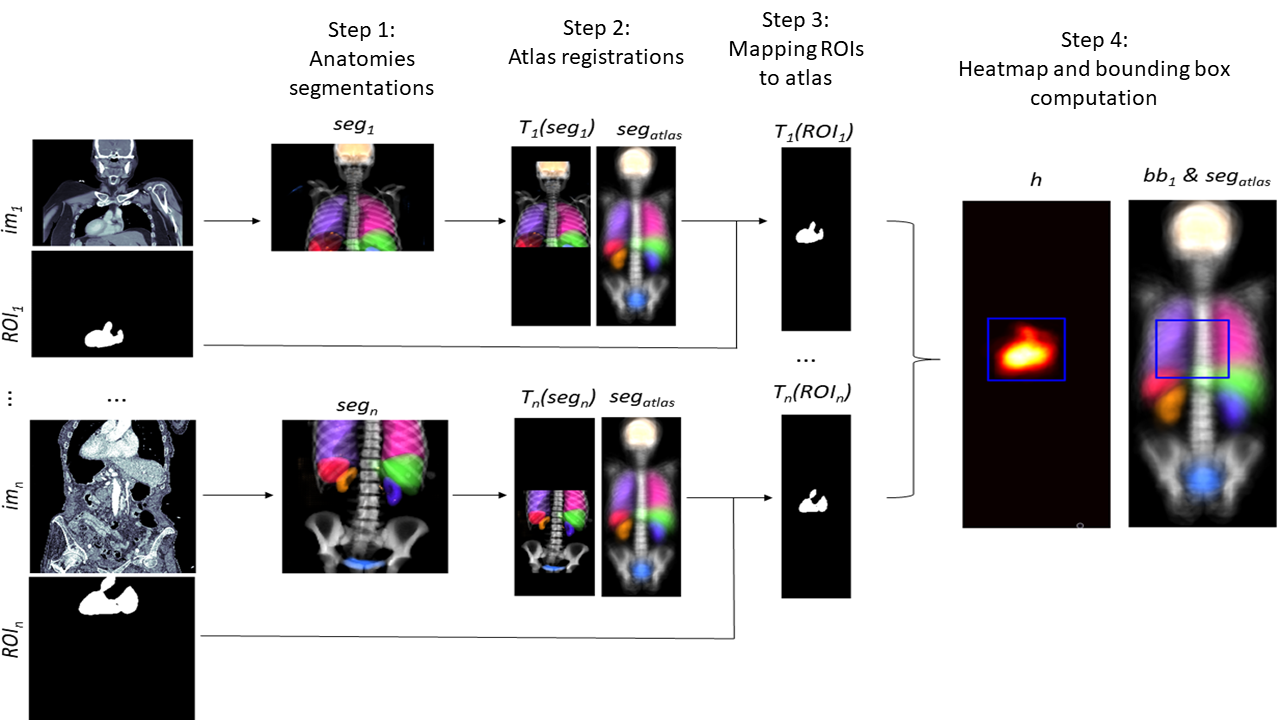}
    \caption{Workflow of computing bounding boxes from labelled CT images. Step 1 and 2 compute the anatomy segmentations and the atlas registration transformation for each scan of the dataset. Step 3 applies these transformations to the regions of interest to map them all into the same coordinate system. Step 4 overlays these to a heatmap and computes the bounding box.}
    \label{fig:bounding_box_computation}
\end{figure}

One way to create anatomical regions is to create or edit the bounding boxes in the atlas coordinate system guided by expert knowledge. Another way is to use pairs of images and segmentations of the region of interest, which is performed the following way. For each image $im_i$ with corresponding region of interest segmentation $ROI_i$ the registration transformation $T_i$ is computed as detailed in Section \ref{atlas_registration}. This transformation is applied to $ROI_i$. The collection of transformed region of interest segmentations $(T_i(ROI_i))_{i=1}^n$ now reveals in which parts of the atlases coordinate system region of interest can and cannot occur. To extract one our multiple bounding boxes from this information we propose to simply average these masks by computing the heatmap
\[ h = \frac1n\sum_{i=1}^n T_i(ROI_i).\]
The bounding boxes can be computed such that they contain all coordinates $(x,y,z)$ for which the heatmaps value is above a certain threshold.  In case all segmentation labels are clean one can use 0 as a threshold, otherwise a larger threshold can be used. Our implementation computes the bounding box for each connected components of the thresholded heatmap $h$ and merges overlapping ones. To account for anatomical variations and imperfections in the atlas registrations we increased the resulting bounding boxes by 1cm in each direction.
The corresponding workflow is visualized in Figure \ref{fig:bounding_box_computation}.

Our code is available at https://github.com/ThomasBudd/ctarr and is delivered with bounding boxes for the anatomical regions of the kidneys, brain, colon, gallbladder, heart, liver, lungs, pancreas, spine, spleen, stomach, and urinary bladder. The creation of these boxes is described in the following section.

\section{Experiments}

\subsection{Image registration}
We use the Totalsegmentator dataset to compare our proposed segmentation and atlas registration algorithm with alternative methods. First, to assess the influence of the errors of the segmentation network onto the registration result, we ran the registration algorithm on this dataset again using the manual ground truth segmentations instead of the segmentation produced by the CNN. Second, to compare our registration pipeline with established state-of-the-art algorithms that perform directly on CT images, we use the affine registration of the ANTs framework\footnote{https://antspy.readthedocs.io/en/latest/registration.html}. To perform image-based atlas registration we needed to create a CT image atlas as the atlas of our registration method is only given as a set of segmentations. For this we used the scan with the longest z axis and registered it and its segmentation with our registration method to the atlas and verified the results visually (see Figure \ref{fig:ants_failiures}). Lastly, we set up our proposed method. Since the original segmentation network was trained on the Totalsegmentator dataset, we trained the network again in four-fold cross-validation and collected only the predictions of scans that were not present in the training data to prevent a bias due to overfitting. We quantified the results in terms of DSC between the manual ground truth anatomy segmentations and NCC between the CT image atlas and the warped moving image. 

\subsection{Anatomical region cropping}
Next, we aimed at testing the utilities of the cropping pipeline. For this, we used the Totalsegmentator dataset to create a set of bounding boxes for various anatomical regions without considering other data sources. As the labels of the dataset are noisy, we could not threshold the heatmap $h$ (see Section \ref{region_computing}) with a value of 0, but instead had to choose larger values to prevent single outliers (see Appendix A) from greatly increasing the bounding boxes. The thresholds used for each region is listed in the supplementary Table \ref{tab:thresholds}. The data was obtained by using random scans from different scanners of the picture archiving system of a university hospital. In this case it is reasonable to assume that the majority of the segmented anatomies were healthy.\\
The cropping pipeline was tested on a total of n=1131 scans from public segmentation challenges namely the liver, lung, pancreas, spleen and colon dataset of the medical segmentation decathlon (MSD, \cite{medicaldecathlon}) and the kits23 challenge. In contrast to the Totalsegmentator data, these datasets contain scans of pathological organs (except for the spleen dataset).\\
As a measure for the sensitivity of the method, we computed the percentage of preserved foreground voxels after cropping. The specificity was captured by computing the percentage of foreground voxels contained in the image before and after cropping. To motivate the usage in reducing the inference time of image segmentation pipelines we further measured (1) the time it takes to execute the cropping pipeline, (2) the execution speed of inference of an nnU-Net style segmentation algorithm with and (3) without cropping the images using our proposed method. More technical detail on these experiments can be found in the Appendix B. It should be noted that we did not train an nnU-Net style network for this as this would have required additional training datasets for each segmentation problem. Instead, the inference was simulated by using networks with randomly initialised weights. To check how often the orientation in terms of 90 degrees rotation in xy-plane and flipping of the z-axis was obtained correctly by the pipeline, we manually checked the axial, coronal and sagittal view of each volume after applying our pipeline and computed the percentage of correct orientations.\\

\subsection{Computation of new anatomical regions}
As a last line of experiments we gradually reduced the number of scans from the Totalsegmentator dataset used to compute the bounding boxes and compared those with the MSD pancreas test dataset. This was done to see how many annotated scans are needed to perform the computation of bounding boxes described in Section \ref{region_computing}. As mentioned in Section \ref{region_computing}, we increase the bounding box with a margin of 1cm in each direction by default. To make the bounding boxes obtained by the differing amount of scans comparable, we adapted this additional margin such that the volume of the bounding box equalled the volume of the bounding box created with the full Totalsegmentator dataset.

\section{Results}

\subsection{Image registration}
\begin{figure}[h]
    \centering
    \includegraphics[width=0.8\linewidth]{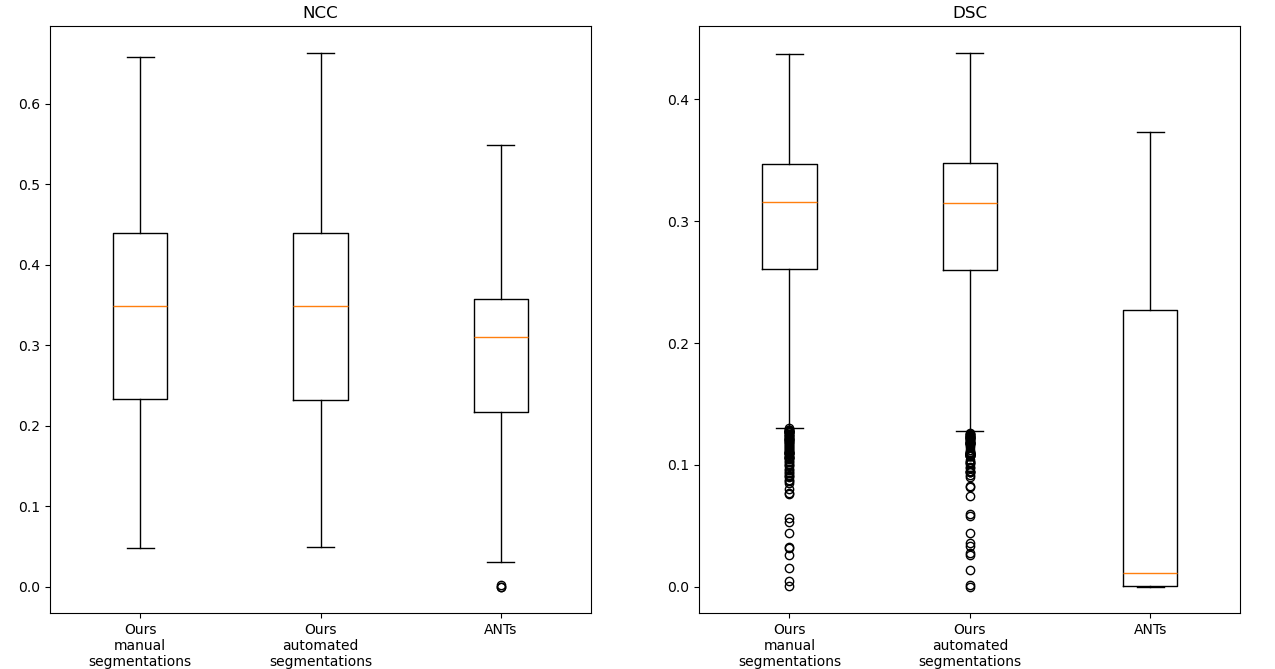}
    \caption{Results of comparing the presented image registration method with the usage of clean segmentation labels and state-of-the-art affine registration methods using CT images instead of segmentations.}
    \label{fig:im_reg_results}
\end{figure}

\begin{figure}[h]
    \centering
    \includegraphics[width=1.0\linewidth]{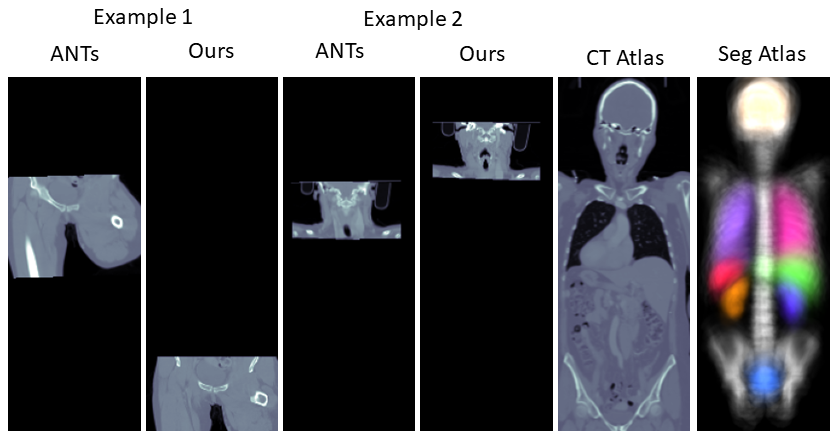}
    \caption{Examples of failures of the traditional image-based registration method.}
    \label{fig:ants_failiures}
\end{figure}

The results of the image registration experiments can be found in Figure \ref{fig:im_reg_results}. 
It can be observed that the results of the registration pipeline using the automated and the manual ground truth labels are almost indistinguishable. While the traditional image-based method maintains moderate NCC values in some cases, it can be observed that the DSC values of this method are very poor. One reason for this is that the method is dependent on the starting point of the iteration and can get stuck in local minima before reaching the desired global minima. Figure \ref{fig:ants_failiures} demonstrates some of such failures.
We present further comparisons via scatter plots in the Supplementary Materials along with visual comparisons.
\subsection{Anatomical region cropping}

\begin{figure}[h]
    \centering
    \includegraphics[width=1.0\linewidth]{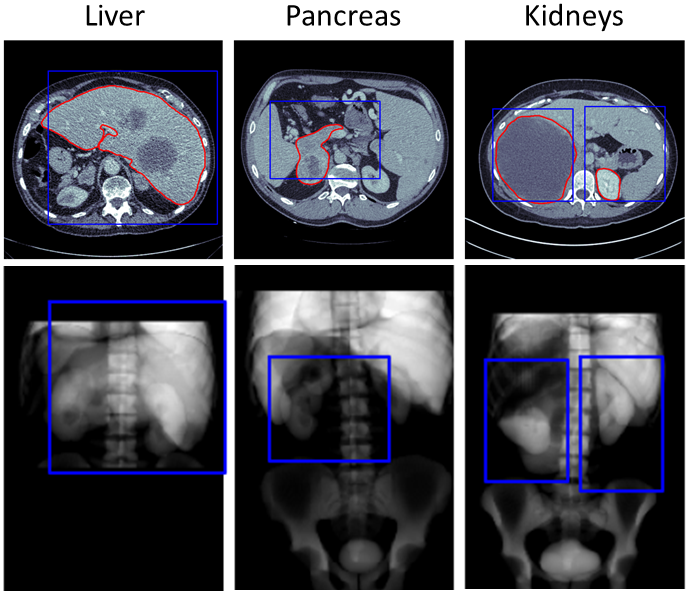}
    \caption{Examples of scan with a loss of ground truth foreground after cropping.}
    \label{fig:outlier_examples}
\end{figure}

The results of testing the cropping pipeline can be found in Table \ref{tab:results}.
\begin{table}[h] 
    \centering
    \caption{Numerical evaluation of the cropping pipeline on the test dataset. The reduction of image voxels was computed by dividing voxels before cropping by the number of voxels after cropping. Computation times were measured using an RTX 6000 Ada GPU with pytorch 2.02 installation.}
    \begin{tabular}{l||l|l|l|l|l|l}
    dataset & Liver  & Lung & Panc. & Spleen & Colon & kits23 \\\hline\hline
    region of & liver \&& lung & panc. \&& spleen & colon & kidneys \&\\
    interest & tumors & cancer & masses & &cancer & masses \\\hline
     num. scans & n=131 & n=63 & n=281 & n=41 & n=126 & n=489 \\\hline
     preserved & 98.91\% & 100\% & 99.07\% & 100\% & 100\% & 97.45\%\\
     foreground & & & & & &\\ \hline
     foreground voxels& 2.28\% & 0.04\% & 0.22\% & 0.44\% & 0.05\% & 0.83\% \\
     before cropping & & & & & &\\ \hline
     foreground voxels& 11.38\% & 0.08\% & 2.08\% & 8.09\% & 0.13\% & 10.3\% \\
     after cropping & & & & & &\\ \hline
     exec. speed & 0.21s & 0.15s & 0.10s & 0.14s & 0.16s & 0.17s \\ 
     cropping & & & & & &\\ \hline
     inference speed & 126.7s & 87.4s & 72.7s & 102.3s & 124.5s & 117.4s\\
     no cropping & & & & & &\\ \hline
     inference speed & 26.5s & 43.4s & 5.7s & 10.2s & 58.1s & 14.0s\\
     cropping & & & & & &\\ \hline
     correct & 99.24\% & 100\% & 100\% & 100\% & 98.41\% & 100\% \\
     orientation & & & & & &\\
    \end{tabular}
    \label{tab:results}
\end{table}
It can be observed that all foreground was preserved in three of the six tasks, and only 0.93-2.55\% of the foreground was lost in the remaining tasks, while the percentage of foreground voxels in the images after cropping was increased by a factor of 1.98-8.13. Figure \ref{fig:outlier_examples} shows examples of scans from each of those three tasks with a loss of ground truth foreground after cropping. Visual analysis revealed that the foreground that got lost during the cropping was always in close proximity to the margin of the bounding boxes. Figure \ref{fig:outlier_examples} also shows that all three scans are still reasonably well aligned with the atlas despite the fact that the pathological masses present in the images caused a distortion in the neighboring anatomies. Table \ref{tab:results} shows that the proposed cropping pipeline took an average of 0.10-0.21s to compute on our RTX 6000 ADA GPU using pytorch 2.02, while reducing the inference computing time of a nnU-Net style segmentation algorithm by 44.0-100.2s. Last, the orientation of the scans was correctly determined in all except one scan in the MSD Liver and two scans in the MSD Colon dataset. The one scan from the MSD Liver dataset was flipped on the z-axis compared to the atlas as the segmentation algorithm labelled the vertebrae in the lower abdomen as vertebrae in the neck. We found that both scans from the MSD Colon dataset with wrong orientation were caused by a corruption of the image files. In both cases, we observed that some bottom slices were placed on the top of the stack of slices instead of the bottom, which also caused a flip of the z-axis when being compared to the atlas.

\subsection{Computation of new anatomical regions}

In the end we evaluated the bounding boxes of the pancreas obtained with the differing amount of CT image and segmentation pairs. The results can be found in Table \ref{tab:pancreas_experiment}. It can be noted that the sensitivity remains very stable even when using as little data as n=25 scans.

\begin{table}[h] 
    \centering
    \caption{Change of sensitivity when decreasing the dataset size for locating the anatomical region of the pancreas. The bounding box size was fixed to the size of the bounding box when using the full dataset.}
    \begin{tabular}{c|c|c|c|c|c|c|c|c}
$n$ & 807 & 403 & 201 & 100 & 50 & 25 & 12 & 6\\ \hline
sens & 99.07 & 99.24 & 99.20 & 99.31 & 99.02 & 98.99 & 98.66 & 98.89
    \end{tabular}
    \label{tab:pancreas_experiment}
\end{table}

\section{Discussion}

The fact that CT images naturally vary in the anatomical context they contain can make automated processing of them more challenging. In this manuscript we proposed the first general purpose pipeline for identifying anatomical regions on CT images. We demonstrated how our pipeline can be adapted to any anatomical region by simply defining a new bounding box in the coordinate system of the atlas. We also showed that our method is robust, fast and can be used to drastically improve the inference speed of segmentation pipelines. This was done using over 1100 scans from six different publicly available datasets. We also suggested a novel registration pipeline based on automatically generated segmentation masks instead of using CT images directly. While this seems like an additional step that might introduce accumulated errors, we demonstrated how the usage of such segmentation masks instead prevents registration methods from getting stuck in local minima. The fact that the registration results using the automated and manual segmentations were extremely similar implies that further improvement of the segmentation model might not lead to an improved registration performance. Further, we demonstrated that potentially only very few labeled examples might be needed to compute new anatomical regions automatically.\\
Our method has room for improvement. Our results showed that inferring bounding boxes from healthy regions of interest and applying it to unhealthy patients might lead to a loss of information when applying the cropping pipelines. This is naturally to be expected as unhealthy organs can be enlarged and might exceed the boundaries of healthy organs. In consequence, more anatomical regions from unhealthy subjects should be computed and added to the pipeline to ensure no loss of information in these cases. As an alternative, an additional cropping margin can be used when using bounding boxes computed from healthy organs for scans with unhealthy organs. Second, the current version of the algorithm is not applicable for CT scans of the legs, as our method is not capable of segmenting anatomies in the legs. The current release of the Totalsegmentator algorithm does feature the segmentation of bones in the legs, however the corresponding labels have not been released in the publicly available dataset yet. Third, while the pipeline was able to infer the orientation of the scan in terms of z-axis flipping and xy-plane rotations, the user still has to know on whether the input images are axial, sagittal, or coronal images. This knowledge can be difficult to obtain especially when using nifti scans with isotropic resolution or false information was added to dicom headers. Fourth, to prevent the orientation alignment errors it is desirable to check the automated anatomical segmentations for plausibility. This might allow raising warnings for scans with labelling errors in the automated segmentation, or corrupted image files as observed in the MSD colon dataset. Finally, we believe that it would be valuable to extend the library for the usage of MRI scans as the Totalsegmentator dataset was recently extended by \cite{totalsegmentatorMRI} for the usage in MRI.\\
For the future, we believe it to be valuable to test the proposed pipeline as a preprocessing step for CT image analysis tasks. For example, the pipeline can be used as a first step in image registration problems where two scans of the same patient are present. In this scenario, CTARR reduces the workload and serves as a pre-registration tool by providing two crops of the same anatomical region. Further, the pipeline should be tested for image classification pipelines targeting cancers with complex distributions like ovarian cancer that is spread across the abdomen. The proposed pipeline can help reducing a dataset of CT scans to image crops of similar size showing the same anatomical context. This might be a key pre-processing step to avoid overfitting and reduce the training and inference burden.\\

\section{Conclusion}

We presented a novel pipeline for cropping anatomical regions on CT images based on automated CNN-based anatomy segmentations and traditional image registration techniques to a predefined atlas. We demonstrated that the pipeline is fast, robust can be adapted easily to identify new anatomical regions.


\acks{We acknowledge funding and support from Cancer Research UK and the Cancer Research UK Cambridge Centre [CTRQQR-2021-100012], The Mark Foundation for Cancer Research [RG95043], GE HealthCare, and the CRUK National Cancer Imaging Translational Accelerator (NCITA) [A27066]. Additional support was also provided by the National Institute of Health Research (NIHR) Cambridge Biomedical Research Centre [NIHR203312] and EPSRC Tier-2 capital grant [EP/P020259/1]. MCO received support from the Joseph Mitchell Fund. The funders had no role in study design, data collection and analysis, decision to publish, or preparation of the manuscript.}

%
\ethics{The work follows appropriate ethical standards in conducting research and writing the manuscript, following all applicable laws and regulations regarding treatment of animals or human subjects.}

\coi{The conflicts of interest have not been entered yet.}

\data{All data used in this study is publicly available. The Totalsegmentator data can be found at https://zenodo.org/records/10047292. The Medical Decathlon data is available via http://medicaldecathlon.com/. The kits21 data can be accessed via the official github repository: https://github.com/neheller/kits21}

\bibliography{sample}


\clearpage
\appendix
\section{Outlier examples in Totalsegmentator}
\begin{figure}[h]
    \centering
    \includegraphics[width=1.0\linewidth]{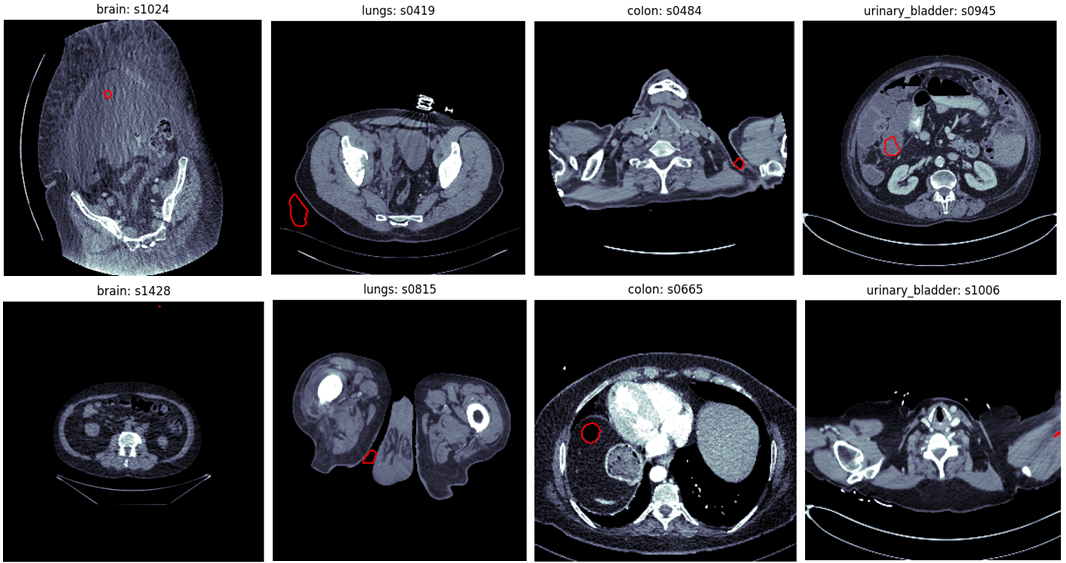}
    \caption{Examples of implausible annotations on the Totalsegmentator dataset. These errors are most likely a result of false positive errors of the segmentation network that were missed by the annotator during the refinement process. The heading of each image displays the label of the red contour and the ID of the scan.}
    \label{fig:totalsegmentator_errors}
\end{figure}

This section briefly demonstrates some outliers far from the actual anatomical region that are introduced by automated segmentation methods. The examples were taken from the Totalsegmentator dataset \cite{totalsegmentator}. The annotations of this dataset were created by first training the nnU-Net framework \cite{nnunet} on little manually created annotations, evaluating the model on unannotated data and correcting the predictions manually. We found that some of the annotations in the Totalsegmentator dataset contained mistakes which are most likely caused by the annotator missing errors of the nnU-Net segmentation. Examples of such can be found in Figure \ref{fig:totalsegmentator_errors}. All errors presented are far away from the anatomical region where the organs are actually located. This might be a reason why the annotators have missed these false positive predictions.

\begin{table}[h] 
    \centering
    \caption{Thresholds applied to infer bounding boxes on the Totalsegmentator dataset. Noise in the Totalsegmentator labels cause too large bounding boxes when using a threshold of 0. The thresholds were obtained by increasing the threshold until the visual inspection of the bounding box in the atlas coordinate system was satisfying.}
    \begin{tabular}{c | c}
    Region & threshold \\ \hline
    spine & 0.03 \\
    brain & 0.01 \\
    lungs & 0.04 \\
    heart & 0.03 \\
    spleen & 0.01 \\
    kidneys & 0.01 \\
    urinary bladder & 0.01 \\
    colon & 0.03 \\
    gallbladder & 0.01 \\
    pancreas & 0.01 \\
    stomach & 0.01 \\
    liver & 0.035 \\
    \end{tabular}
    \label{tab:thresholds}
\end{table}

\section{Hyper-parameters of image augmentation}
The hyper-paramters of the image augmentation can be found in Table \ref{tab:hyper_parameters_aug}. Rotations and grid scaling were applied only in xy plane. Rotations were performed around the image center once using interpolation (magnitude=[-45, 45] degrees) and in a interpolation-free manner (90-, 180-, or 270-degrees). The standard deviation of the additive Gaussian noise was computed by respecting the standard deviation of the image patch $\sigma(im)$. The magnitude of the Gaussian Blurring is given in number of voxels with full width at half maximum.
\begin{table}[h] 
    \centering
    \caption{Hyper-paramters of the image augmentations.}
    \begin{tabular}{c | c | c}
    Operation & Probability & Magnitude\\ \hline
    Rotation & 0.5 & [-45, 45] \\
    Rotation & 0.75 & \{90, 180, 270\}\\
    Grid scaling & 0.5 & [0.7, 1.4] \\
    Flipping (z only) & 0.5 & \\
    Addative Gaussian Noise & 0.5 & $\sigma(im) \times [0, 0.25]$ \\
    Gaussian Blurring & 0.5 & [0, 5] \\
    \end{tabular}
    \label{tab:hyper_parameters_aug}
\end{table}

\section{Additional results image registration}
\begin{figure}[h]
    \centering
    \includegraphics[width=1.0\linewidth]{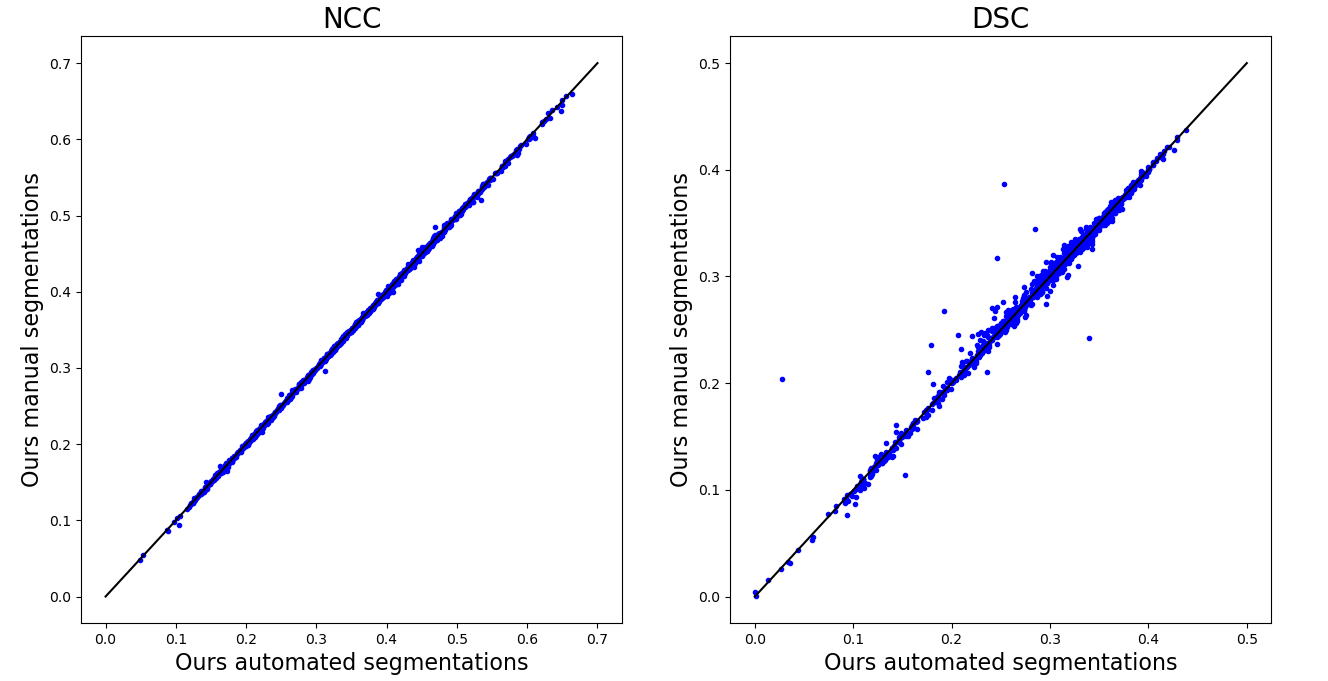}
    \includegraphics[width=1.0\linewidth]{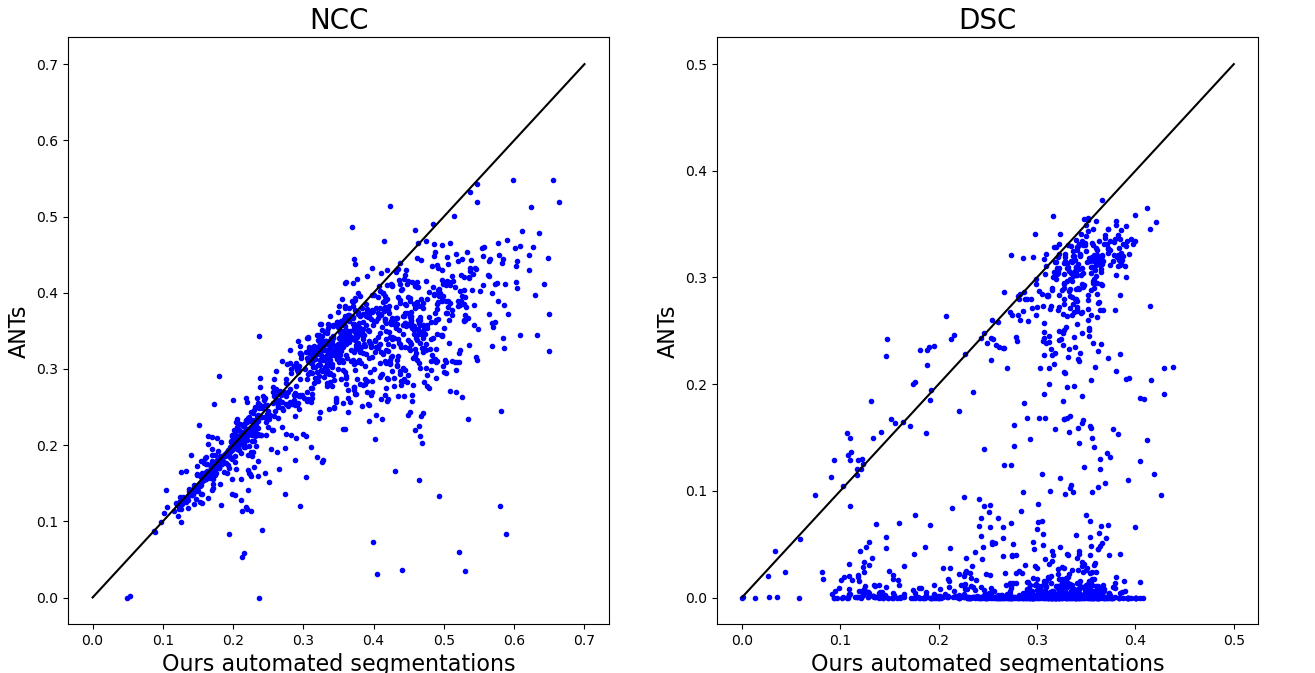}
    \caption{Scatter plot for a case by case comparison of the different registration methods.}
    \label{fig:scatter_plot_im_reg}
\end{figure}

\begin{figure}[h]
    \centering
    \includegraphics[width=1.0\linewidth]{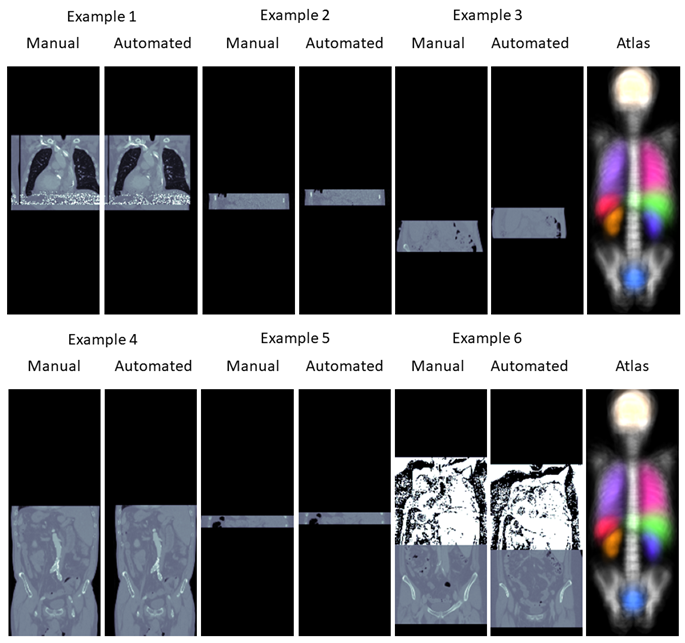}
    \caption{Examples for which the registration achieved a higher DSC performance when using the manual ground truth segmentations instead of the automated segmentations.}
    \label{fig:gt_vs_ours_outliers}
\end{figure}

Figure \ref{fig:scatter_plot_im_reg} presents scatter plots to compare the different registration methods. 
It can be observed that the segmentation-based approaches result in very similar results in terms of NCC and that only few outliers can be detected in terms of DSC. The most dominant six cases in which a higher DSC was achieved by using the manual segmentation instead of the automated segmentation are displayed in Figure \ref{fig:gt_vs_ours_outliers}.
Overall it can be observed how the classical image-based method frequently fails in aligning the CT image with the atlas. 

\section{Simulation of segmentation network inference}
To simulate the computational burden of segmentation network inference we created a inference pipeline similar to the one suggested by nnU-Net \cite{nnunet}. In a first step we resized the images to a common voxel spacing. While nnU-Net suggests a different voxel spacing per dataset, we used a common spacing of 3mm in z and 0.75mm in xy direction to simplify the experiments. Next, we used a 3d U-Net architecture as suggested by nnU-Net. We used a five stage network where the first two stages used only inplane convolutions to account for the anisotropic resolution. The network was evaluated with the Sliding Window algorithm using a patch size of 64 x 160 x 160 voxels. As in nnU-Net, this was repeated applied total of 40 times to simulate the test time augmentation (evaluating with eight configurations of flipping over the z, x, and y axes) and the ensembling (evaluate five networks trained in cross-validation). Finally, the predictions were interpolated again to the original image spacing. In our experiments we used the GPU interpolation operations implemented in pytorch to accelerate the inference. We used an NVIDIA 6000 Ada GPU and pytorch v2.02 for all experiments.
\end{document}